\documentclass[runningheads]{llncs}
\usepackage{times}  
\usepackage{helvet} 
\usepackage{courier}  
\usepackage[hyphens]{url}  
\usepackage{graphicx} 
\usepackage{multirow}
\usepackage{arydshln}
\usepackage{xcolor}
\usepackage{graphicx}  

\begin{document}

\title{Constructing Artificial Data for Fine-tuning for Low-Resource Biomedical Text Tagging} 
\subtitle{with Applications in PICO Annotation}
\titlerunning{Constructing Artificial Data for Low-Resource Biomedical Text Tagging}
\author{Gaurav Singh\inst{1},
Zahra Sabetsarvestani\inst{2},
John Shawe-Taylor\inst{1},
James Thomas\inst{1}\\
}
\authorrunning{G. Singh et al.}
\institute{ University College London\\
\email{\{g.singh\}@cs.ucl.ac.uk} \\
\email{\{j.shawe-taylor, james.thomas\}@ucl.ac.uk} \and
AIG\\
\email{\{zahra.sabetsarvestani\}@aig.com}}

\maketitle

\begin{abstract}
    Biomedical text tagging systems are plagued by the dearth of labeled training data. There have been recent attempts at using pre-trained encoders to deal with this issue. Pre-trained encoder provides representation of the input text which is then fed to task-specific layers for classification. The entire network is fine-tuned on the labeled data from the target task. Unfortunately, a low-resource biomedical task often has too few labeled instances for satisfactory fine-tuning. Also, if the label space is large, it contains few or no labeled instances for majority of the labels. Most biomedical tagging systems treat labels as indexes, ignoring the fact that these labels are often concepts expressed in natural language e.g. `Appearance of lesion on brain imaging'. To address these issues,  we propose constructing extra labeled instances using label-text (i.e. label's name) as input for the corresponding label-index (i.e. label's index). In fact, we propose a number of strategies for manufacturing multiple artificial labeled instances from a single label. The network is then fine-tuned on a combination of real and these newly constructed artificial labeled instances. We evaluate the proposed approach on an important low-resource biomedical task called \textit{PICO annotation}, which requires tagging raw text describing clinical trials with labels corresponding to different aspects of the trial i.e. PICO (Population, Intervention/Control, Outcome) characteristics of the trial. Our empirical results show that the proposed method achieves a new state-of-the-art performance for PICO annotation with very significant improvements over competitive baselines. 
\keywords{Biomedical Text Tagging \and PICO Annotation \and Artificial Data \and Transfer Learning.}
\end{abstract}


\section{Introduction}
Biomedical text is often associated to terms that are drawn from a pre-defined vocabulary and indicate the information contained in the text. As these terms are useful for extracting relevant literature,  biomedical text tagging has remained an important problem in BioNLP \cite{zweigenbaum2007frontiers,demner2016aspiring} and continues to be an active area of research \cite{singh2018structured,singh2017neural}. Currently, a standard text tagging model consists of an encoder that generates a context-aware vector representation for the input text \cite{yang2016hierarchical,adhikari2019docbert,zhou2015c}, which is then fed to a 2-layered neural network or multi-layered perceptron that classifies the given text into different labels. The encoder is generally based on CNN \cite{lai2015recurrent,lee2016sequential,yang2016hierarchical}, RNN \cite{lee2016sequential}, or (more recently) multi-head self-attention network e.g. Transformer \cite{adhikari2019docbert}. This works well for tasks that have a lot of labeled data to train these models. 
\begin{figure}
    \centering
    \includegraphics[scale=0.35]{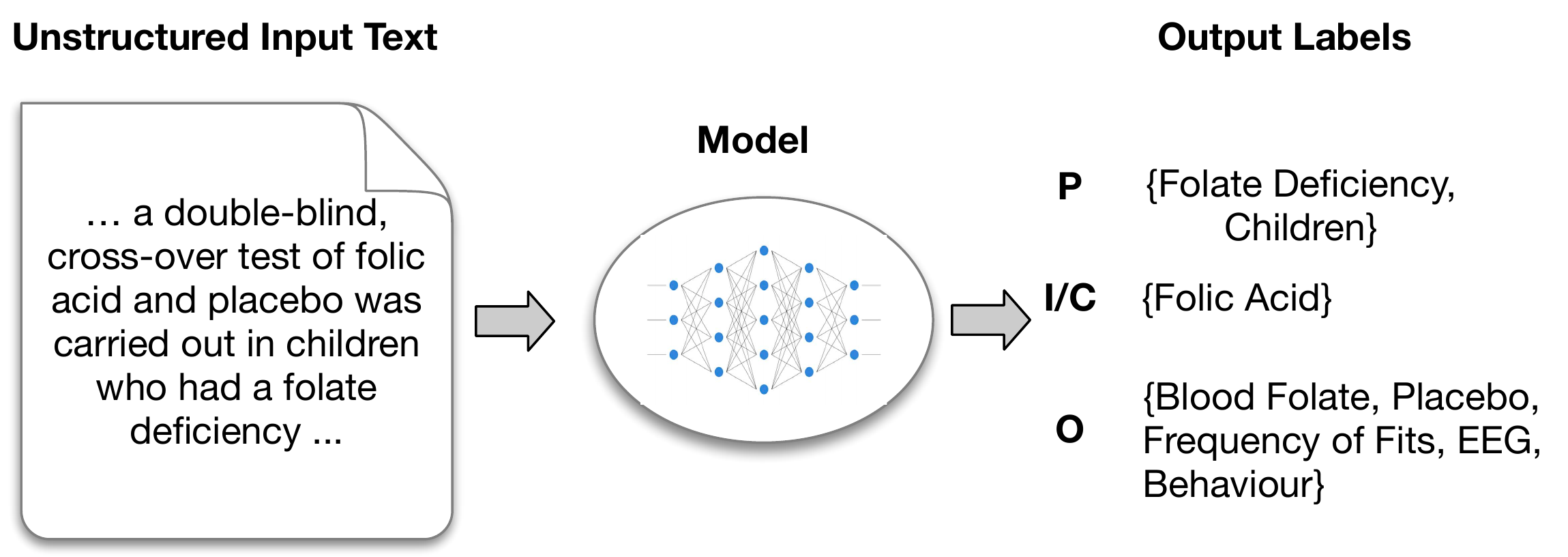}
    \caption{An illustration of the PICO annotation task. The model receives input text describing a clinical trial and applies concepts from a pre-defined vocabulary corresponding to Population, Intervention and Outcome characteristics of the trial.}
    \vspace{-1em}
    \label{fig:my_label}
\end{figure}


Unfortunately, lack of labeled data is a common issue with many biomedical tagging tasks. We focus on one such low-resource biomedical text tagging task called \textit{PICO} annotation \cite{singh2017neural}. It requires applying distinct sets of concepts that describe complementary clinically salient aspects of the underlying trial: the population enrolled, the interventions administered and the outcomes measured, i.e. the PICO elements. Currently, these PICO annotations are obtained using manual effort, and it can take years to get just a few thousand documents annotated. Not surprisingly, there are not enough PICO annotations available to train a sophisticated classification model based on deep learning. 

Recently, there have been attempts to address this issue of low-resource biomedical text tagging using transfer learning \cite{rios2019neural,zeng2019automatic}. These methods begin by training a (large) deep neural network model on a high-resource biomedical task followed by fine-tuning on the low-resource target task. One such work from Zeng et al. \cite{zeng2019automatic} proposed training a deep CNN on the source task of applying MeSH tags to biomedical abstracts, and then used this pre-trained network for applying ICD-9 codes to clinical text after fine-tuning. They used a freely available repository of biomedical abstracts on Pubmed, which are annotated with medical concepts called MeSH tags (i.e. Medical Subject Heading) for source task. In a similar work, Rios et al. \cite{rios2019neural} focused on a different target task and experimented with ablations of network architectures. There have also been attempts in the past to solve PICO annotation using concept embeddings pre-trained using deepwalk on the vocabulary graph \cite{singh2017neural}.  
 
One drawback of above methods is that they treat labels as indexes, despite the fact that these labels are concepts expressed in natural language e.g. `Finding Relating To Institutionalization'. After training on the source task, the encoder is retained, while the task specific deeper layers are replaced with new layers, and the entire network is fine-tuned. Consequently, task-specific layers have to be re-trained from the very beginning on a low-resource biomedical task. This poses problems, not only because there is insufficient labeled data for fine-tuning, but there are few or no labeled instances for a large number of labels. 

To address this, we propose constructing additional artificial instances for fine-tuning by using the label-text as an input for the corresponding label-index. We propose various strategies to construct artificial instances using label-text, which are then used in combination with real instances for fine-tuning. This is based on the hypothesis that an encoder pre-trained on a related biomedical source task can be used to extract good quality representations for the label-text, where these label-texts are concepts  expressed in natural language. 


To summarize the contribution of this work: 

     \begin{itemize}
         \item To address the scarcity of labeled data for PICO annotation, we construct additional  artificial labeled instances for fine-tuning using label-texts as inputs for label-indexes.
         \item In order to obtain rich encoder representations for label-text, we pre-train the encoder on a related source task i.e. MeSH annotation.
     \end{itemize}
    

\section{Related Works}
We focus on the problem of low-resource biomedical multi-label text classification using transfer learning. As such there are three broad areas pertinent to our work: (1) multi-label text classification; (2) biomedical text annotation; (3) transfer learning for classification.  We will give an overview of relevant literature in these three areas. 

\subsection{Multi-label Text Classification}
The task of PICO annotation can be seen as an instance of multilabel classification, which has been an active area of research for a long time, and therefore has a rich body of work \cite{mccallum1999multi,elisseeff2001kernel,furnkranz2008multilabel,read2009classifier}. One of the earliest works \cite{mccallum1999multi} in multi-label text classification represented multiple labels comprising a document with a mixture model. They use expectation maximization to estimate the contribution of a label in generating a word in the document. The task is then to identify the most likely mixture of labels required to generate the document. Unfortunately, when the set of labels is prohibitively large, these label mixture models are not practical. Then, a kernel based approach for multilabel classification based on using a large margin ranking system was proposed \cite{elisseeff2002kernel}. In particular, it suggests a SVM like learning system for direct multi-label classification, as opposed to breaking down these into many binary classification problems.  


More recently, there has been a shift towards using deep learning based methods for multilabel classification \cite{nam2014large,liu2017deep,yeh2017learning}. One such work \cite{nam2014large} proposes using a simple neural network approach, which consists of two modules: a neural network that produces label scores, and a label predictor that converts labels scores into classification using threshold. Later, \cite{liu2017deep} focused on the problem of assigning multiple labels to each document, where the labels are drawn from an extremely large collection. They propose using a CNN with dynamic max-pooling to obtain document representation, which is then fed to an output layer of the size of label vocabulary, with sigmoid activation and binary cross-entropy loss. In \cite{yeh2017learning}, they propose learning a joint feature and label embedding space by combining DNN architectures of canonical correlation analysis and autoencoder, coupled with correlation aware loss function. This neural network model is referred to as Canonical Correlated AutoEncoder (C2AE).

\begin{figure*}[hbt]
    \centering
    \includegraphics[scale=0.28]{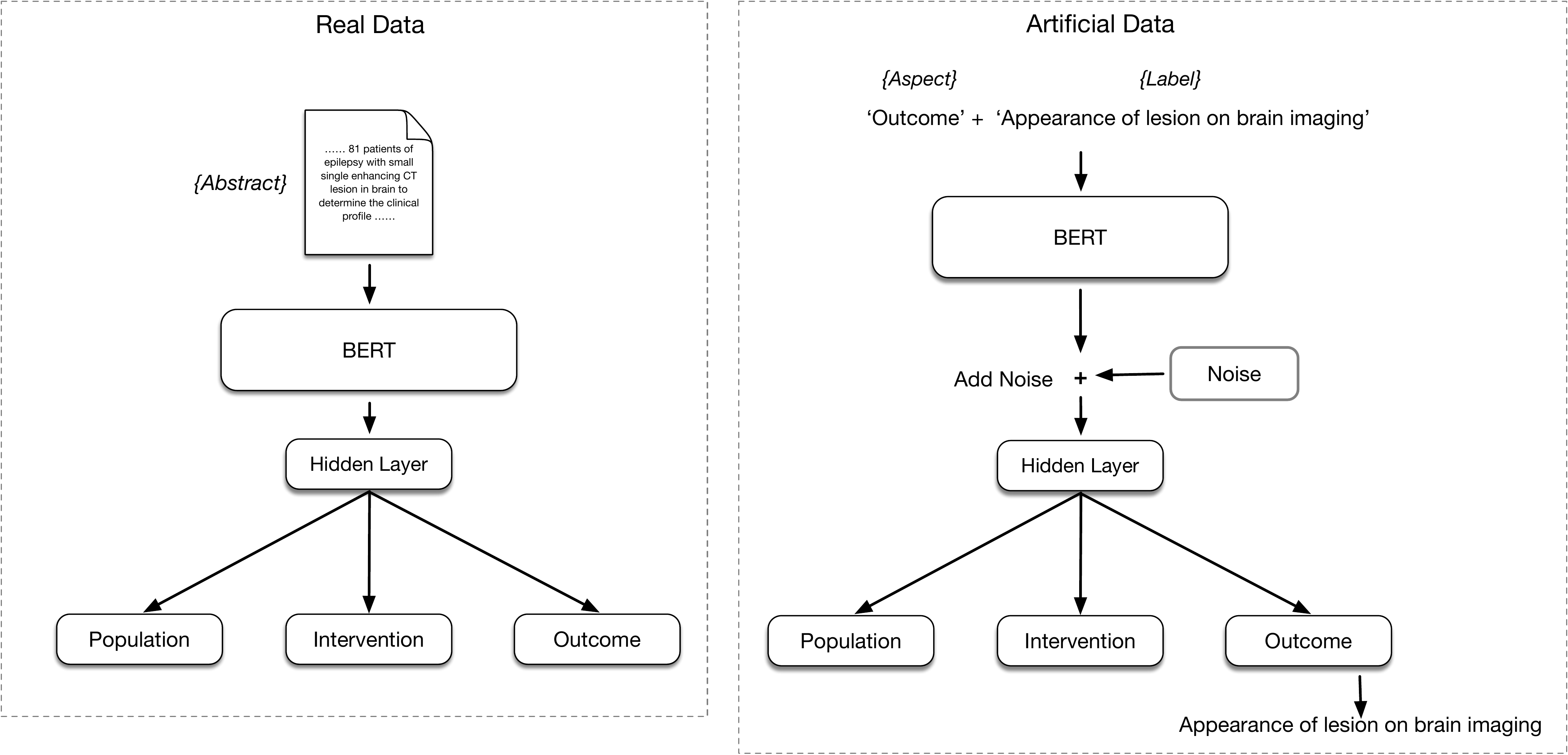}
    \caption{A pictorial representation of the model architecture and training process. \textbf{Left}: The model takes as input an unstructured piece of text e.g. abstract, and applies the assigned labels. \textbf{Right}:  To address scarcity of labeled data for fine-tuning, we construct artificial inputs for labels using the label-text as input for the label-index.}
    \label{fig:arch}
\end{figure*}
\subsection{Biomedical Text Annotation}
Biomedical text annotation is a very broad and active area of research \cite{demner2016aspiring,zweigenbaum2007frontiers,rios2019neural,singh2017neural,zeng2019automatic}, e.g. PICO annotations and MeSH tagging. The authors in Singh \emph{et al.} \cite{singh2017neural} focused on the problem of PICO annotation under the constraint of sparse labelled data. Their approach is based on generating a list of candidate concepts and then filtering this list using a deep neural network. The candidates concepts for a given abstract are generated using a biomedical software called Metamap, which uses various heuristic, rules and IR-based techniques to generate a list of concepts that might be contained in a piece of text. 

Another work \cite{singh2018structured} focused on the similar task of MeSH tagging such that the tags are drawn from a tree-structured vocabulary. They treat biomedical text tagging as an instance of seq-to-seq modelling so that the input text sequence is encoded into a vector representation, which is then decoded into a sequence corresponding to the path of a label from the root of the Ontology. A similar work \cite{mullenbach2018explainable} on applying ICD (diagnostic) codes to clinical text broke down the multilabel classification problem into many binary classification problems. They use a CNN to obtain feature maps of the input text, and then apply per-label attention to obtain a contextual representation, which is then fed to a single-neuron output layer with sigmoid activation. They focus on explaining predictions using the attention distribution over the input text and share the feature maps generated using CNN across different labels. 

\subsection{Transfer Learning for Text Classification}
There have been various attempts to use transfer learning for text classification in the past \cite{do2006transfer,dai2007transferring,pan2009survey,dai2009translated}, but with very limited success. Some of these earlier methods were based on inferring parameters of the classification model from dataset statistics \cite{do2006transfer}, thus eliminating the need to train on every task. While this is an interesting approach, these statistics are often unable to transfer anything more than general characteristics of the model that are available e.g. mean or variance. In a different line of work  \cite{pan2008transfer}, authors proposed an algorithm based on projecting both source and target domain onto the same latent space before classification. 

Lately, the popularity of deep learning models for solving various NLP tasks has multiplied manifolds.  As a matter of fact, the two works closest to us from Rios \emph{et al.} \cite{rios2019neural} and Zeng \emph{et al.} \cite{zeng2019automatic} are based on CNNs. Both the methods begin by pre-training a CNN on tagging abstracts with MeSH terms, which is then used for applying  ICD-9 (disease codes) codes to unstructured EMRs.  More recently, significant success has been achieved on variety of NLP tasks \cite{adhikari2019docbert,rios2019neural,zeng2019automatic} by using pre-trained bidirectional encoders. Hence, instead of CNN, we use one such pre-trained bidirectional encoder in this work. 

\section{Method}
In this section, we begin with a brief description of the model architecture used in PICO annotation. This includes a description of the encoder architecture used for both, source and target task. Then, we give details of the source task (i.e. MeSH annotation) used for pre-training the encoder. This is followed by a discussion on the target task (i.e. PICO annotation) where we provide motivations for constructing artificial instances. We also describe our proposed strategies for constructing artificial instances to be used in fine-tuning. Finally, we briefly mention our fine-tuning strategy that uses a mixture of real and artificial data.  

\label{sec:method}
\subsection{Model}
We provide an overview of the model architecture used for PICO annotation in Figure \ref{fig:arch}. The model consists of a bidirectional encoder (i.e. Transformer encoder), followed by a shared hidden layer and three task-specific output layers - one each for concepts corresponding to aspects P, I/C and O. 

The encoder provides contextual representation of the input text (e.g. abstract), which is then passed through the shared hidden layers, and then to the task-specific output layers. We append a [CLS] token to the beginning of the input text, and the vector representation of this token is taken to be the vector representation of the input text. The task specific output layers then classify the text into pre-defined vocabulary of PICO concepts. The shared hidden layer is used to leverage correlations between different P, I/C and O concepts, for example, if the trial Population is `\textit{Male}' then Outcome will not generally be `\textit{Breast Cancer}'.

\subsubsection{Encoder.}
We use the encoder of the Transformer \cite{vaswani2017attention} network to generate contextual representations of input text. It consists of repeating blocks and each block consists of two sublayers: multi-head self-attention layer and position-wise convolution layer. There are residual connections and layer normalization after every sublayer. The input is a sequence of word embeddings added with positional embeddings before being fed to the encoder. 

 Recently, it has been shown that pre-trained transformer encoder i.e. BERT \cite{devlin2018bert}, can generate rich representations for input text, leading to superior results over a number of low-resource target NLP tasks. Hence, we  pre-trained BERT on a biomedical task similar to PICO annotation, this task applies MeSH terms to biomedical abstracts available on PubMed\footnote{An online repository of biomedical abstracts.}.



\subsection{Source Task: MeSH Annotation}
We extract over 1 million abstracts from Pubmed along with the annotated MeSH tags, where these MeSH terms correspond to specific characteristics of a biomedical text. The extracted dataset covered the entire vocabulary of MeSH terms with $\approx29$K unique terms. These terms are arranged in the from of a tree-structured hierarchy e.g. \textit{Disease} $\rightarrow$ \textit{Virus Diseases} $\rightarrow$ \textit{Central Nervous System Viral Diseases} $\rightarrow$ \textit{Slow Virus Diseases}. These MeSH terms are similar in functionality to the PICO tags, but they are more coarse and not based on PICO characteristics.  The model consisted of BERT as the encoder, followed by a hidden layer and an output layer of the size of MeSH vocabulary. Afterwards, the hidden layer and the output layer are replaced with new layers. Please note that the vocabulary of PICO and MeSH is completely different.
We trained the model using binary cross-entropy loss for over 45 days on a single GPU (2080) with a minibatch size of 4.
\subsection{Target Task: PICO Annotation}

We use the BERT encoder trained on the MeSH annotation task for PICO annotations. BERT generates vector representation of a document, these representations are then passed through a new hidden layer followed by three separate output layers - one each corresponding to P, I/C and O aspects of the trial - and the entire model is fine-tuned on the PICO annotation task. 


As the task-specific layers are completely replaced, these layers have to be re-trained from scratch on the PICO annotation task. This would not be an issue if there was sufficient labeled data available for effectively fine-tuning these task-specific layers.  Unfortunately, biomedical tasks - such as PICO annotation - have scarce labeled data available for fine-tuning. Also, if the label space is large then a number of labels never appear in the fine-tuning set. 

Interestingly, labels in PICO annotation are often self-explanatory biomedical concepts expressed in natural language e.g. `Finding Relating To Institutionalization'. Since the encoder has been pre-trained on a large corpora of biomedical abstracts, we hypothesize that it can be used to obtain rich contextual representations for these PICO labels. A label-text can then be used as an input for the label-index; in other words, we are constructing artificial instances for labels using the label-text as input and label-index as output. We propose different strategies for constructing artificial labeled instances in this fashion.

\subsubsection{Artificial Instance Construction}
In this section we describe our proposed strategies for constructing artificial inputs for label-indexes. These strategies are based on using label-text to construct artificial inputs. These newly created artificial instances can then be used in combination with real instances for fine-tuning.  Here, we describe these strategies in details:
\begin{itemize}
    \item \textbf{LI:} We refer to the first strategy as LI which stands for \textit{Label} as \textit{Input}. In this strategy, we feed the network with the label-text (e.g. `Finding Relating To Institutionalization') as an input for the label-index. We also append the [CLS] token to the beginning of the resulting string. The encoder generated representation for the label-text is then passed on to the shared hidden layer and then to the (three) output layers for classification. This way we have generated extra labeled instances for labels, which helps in effective fine-tuning of the task-specific layers, especially for labels that are rare. 
    \item \textbf{LIS:} LIS stands for LI + Synonyms. We randomly replace different words in the label-text with their synonyms. More specifically, we randomly select one of the words in the label-text and replace it with its synonym, we then select another word from the label-text and repeat the same process over to generate multiple instances from a single label-text. As an example, `Finding Relating To Institutionalization' $\rightarrow$ `Conclusion Relating To Institutionalization'.  We only replace one word at a time to ensure that semantics of the label-text do not venture too far from the original meaning. These synonyms are extracted from the WordNet corpus.  
    \item \textbf{LISA:} LISA stands for \textit{LIS} + \textit{Aspect}, where Aspect is the P, I/C and O characteristic of the trial. In this strategy, we append one of the aspects to the label-text e.g. `Outcome'+ `Finding Relating To Institutionalization', which is then classified as the label in the output layer corresponding to that specific aspect - we have three output layers corresponding to P, I/C and O aspects. 
    \item \textbf{LISAAS:} LISA+AS stands for \textit{LIS} + \textit{Aspect} + \textit{Auxiliary Sentences}. As opposed to the previous strategy of simply appending the aspect to the label-text, we construct artificial sentences by prepending a pre-defined fixed aspect-based sentence to the label-text. For example, `The population of the trial consists of patients with' + `Diabetes' as an input text for the label 'Diabetes', where the first part is the sentence containing the aspect of the label and last part is the label-text. We prepare a fixed set of three aspect-sentences, one corresponding to each P, I/C and O aspect. 
    \item \textbf{LISAAS + Noise:}  While \textit{LISAAS} is closer to a real abstract than other strategies, it can still bias the model towards expecting the exact sentence structure at test time. We can not possibly construct all possible input texts in this manner to train the network. To address this issue, we add random Gaussian noise to the vector representations of the input text generated by the encoder. The encoder should be able to attend to important bits of information at test time but due to the differences in sentence structure the vector representation might be slightly different. By adding noise we wanted to replicate that scenario during training. We apply layer normalization after adding noise to ensure the norm of the signal remains similar to the real input text. 

\end{itemize}

\subsubsection{Training}
We randomly choose to construct a minibatch from either all real or all artificial instances for the first 50\% epochs during fine-tuning. Afterwards, we continue with only real data until the end.  This training process is followed with all the different strategies for constructing artificial instances. 
\section{Experimental Setup}
In this section, we describe in details  the dataset used for experiments, the baseline models used for comparison and the evaluation setup and the metrics used to assess the performance.
\subsection{Dataset}
\begin{table}
    \centering
    \begin{tabular}{l l}
         samples (clinical trials) &  10137 \\ 
         distinct population concepts & 1832\\ 
         distinct intervention concepts & 1946\\ 
         distinct outcome concepts & 2556\\ 
         population concepts & 15769\\ 
         intervention concepts & 10537\\ 
         outcome concepts & 19547\\ 
    \end{tabular}
    \caption{Dataset statistics.}
    \label{tab:dataset-statsl}
\end{table}
We use a real-world dataset provided by Cochrane\footnote{Cochrane is an international organization that focusses on improving healthcare decisions through evidence: \url{http://www.cochrane.org/}.} which consists of manual annotations applied to biomedical abstracts. More specifically, as we have outlined throughout this paper, trained human annotators have applied tags from a subset of the Unified Medical Language System (UMLS) to free-text summaries of biomedical articles, corresponding to the PICO aspects. We should recall that PICO stands for Population, Intervention/Comparator and Outcomes. Population refers to the characteristics of clinical trial participants (e.g., diabetic males). Interventions are the active treatments being studied (e.g., aspirin); Comparators are baseline or alternative treatments to which these are compared (e.g., placebo) -- the distinction is arbitrary, and hence we collapse I and C. While the Outcomes are the variables measured to assess the efficacy of treatments (e.g., headache severity). 

These annotations are performed by trained human annotators, who attached concept terms corresponding to each PICO aspect of individual trial summaries. 

\subsection{Baselines}
We use three straightforward baselines for comparison, these are: (1) training a CNN-based multitask classifier that directly predicts the output labels based on the input text, (2) BERT pre-trained on \textit{Wikipedia} and \textit{BookCorpus}, and, (3) BERT pre-trained on MeSH tagging task.

The CNN-based model consists of an embeddings layer, which is followed by three parallel convolution layers with filter sizes 1, 3 and 5. Each of these convolution layers are followed by a ReLU activation and max-pooling layer, after which the outputs from these 3 layers are concatenated. We then apply: a dropout layer, dense layer, ReLU activation layer; in that order. Finally, we pass the hidden representation to three different output layers for classification.

The other obvious choice for a baseline is to use BERT pre-trained on BookCorpus \cite{zhu2015aligning} and English Wikipedia, which are the same datasets used in the original paper \cite{devlin2018bert}. The encoder consists of 12 consecutive blocks where each such block consists of 12 multihead attention heads and position-wise convolution layers. The encoder representations are then passed through three separate dense layers where the output of each layer is passed to one of the three output layers. These three output layers classify the input text into corresponding P, I/C and O elements. 

There have been works in the past using CNNs pre-trained on MeSH tagging for separate biomedical annotation tasks. Hence, we decided to pre-train BERT on a dataset of over one million abstracts tagged with MeSH terms. The architecture of the model exactly resembles the one described above. All the baselines are only trained on real data.

\subsection{Evaluation Details}

We divided the dataset into 80/20 for train/test split. We had ground truth annotations for all instances corresponding to PICO aspects of the trial i.e. all texts have been annotated by domain experts with labels from the UMLS vocabulary. The texts here are summaries of each element extracted for previous reviews;  we therefore concatenate these summaries to form continuous piece of text that consists of text spans describing respective elements of the trials.  The exhaustive version of vocabulary used by Cochrane consists of 366,772 concepts, but we restricted the vocabulary to labels that are present in the dataset. We summarize some of these statistics in Table \ref{tab:dataset-statsl}.

\begin{table*}[t]
\begin{center}
  \resizebox{\columnwidth}{!}{
\begin{tabular}{ l | l | c  c  c  c c  c }
 
  Category & Model & Macro-Pr & Macro-Re & Macro-F1 & Micro-Pr & Micro-Re & Micro-F1\\ \hline
   \multirow{8}{*}{Population} & CNN  &  0.031 & 0.027 &  0.027 & 0.614 & 0.506 & 0.555 \\ 
                                         & BERT  & 0.040  & 0.036 &0.037  &0.714  &0.572 & 0.635 \\
                                         & BERT-MeSH & 0.043 & 0.038 & 0.038 &  0.719 & 0.560 & 0.629 \\ 
                                          \cdashline{2-8}
                                         & LI  &  0.048 & 0.044 &  0.044 & 0.726 &  0.604 &  0.659 \\ 
                                         & LIS  & 0.046 & 0.042 & 0.043  &0.733  & 0.596 & 0.658 \\ 
                                         & LISA &0.046&0.042&0.042&\textbf{0.757}& 0.606& \textbf{0.673} \\
                                         & LISAAS &0.048&0.044&0.045&0.740&0.603& 0.664\\
                                         & + Noise &\textbf{0.049}&\textbf{0.046}&\textbf{0.046}&0.737& \textbf{0.613}& 0.670\\\hline
    \multirow{8}{*}{Interventions/Comparator}& CNN & 0.027 & 0.023 & 0.024 & 0.671 &  0.418 & 0.515 \\ 
                                         & BERT & 0.032 & 0.028  & 0.029 & 0.717 & 0.465 &  0.564\\  
                                          & BERT-MeSH & 0.035  & 0.032 &0.032  & 0.729  &0.483  &  0.581 \\ 
                                          \cdashline{2-8}
                                         & LI  &0.043 &0.041 &0.041  & 0.700  &  0.549 & 0.616 \\ 
                                         & LIS  & 0.044 & 0.040 & 0.041 &  \textbf{0.753} & 0.528 & 0.621 \\ 
                                         & LISA &0.042&0.037&0.038& 0.743& 0.526& 0.616\\
                                         & LISAAS &0.044& 0.041&0.041&0.748& 0.540& 0.627\\
                                         & + Noise &\textbf{0.045}&\textbf{0.043}&\textbf{0.042}&0.740&\textbf{0.550}&\textbf{0.631} \\\hline
    \multirow{8}{*}{Outcomes}   & CNN & 0.037 & 0.030 & 0.030 & 0.530 & 0.333 & 0.409 \\ 
                                         & BERT &0.043  & 0.035 &  0.036 &0.603  &0.398 & 0.480 \\
                                          & BERT-MeSH  &0.048 & 0.039 & 0.041 &0.612 & 0.412 &  0.492\\
                                          \cdashline{2-8}
                                         & LI   &0.052& 0.046& 0.046& 0.611& \textbf{0.468} & \textbf{0.530}\\
                                         & LIS   &0.052 & 0.044& 0.045 &0.621& 0.442&  0.517\\
                                         & LISA &0.050&0.042&0.043&\textbf{0.629}&0.442&  0.519\\
                                         & LISAAS &\textbf{0.054}&0.046&0.047&0.620&0.453& 0.523\\
                                         & + Noise &0.053& \textbf{0.047}&\textbf{0.047}&0.620&0.455&0.525 \\\hline
  \end{tabular} 
}
\end{center}
\caption{Precisions, recalls and f1 measures realized by different models on the respective PICO elements. Best result for each element and metric is {\bf bolded}. CNN refers to the CNN-based multitask classifier, BERT refers to the transformer encoder pre-trained using BookCorpus and English Wikipedia in the original paper \cite{devlin2018bert}, while BERT-MeSH refers to the transformer encoder pre-trained on MeSH tagging task. Please note that all methods used the same subword tokenizer provided by BERT. Also, all baselines were trained on only real data.}
\vspace{-1em}

\label{tab:tab1}
\end{table*}

We performed all of the hyper-parameter tuning via nested-validation. More specifically, we separated out 20\% of training data for validation, and kept it aside for tuning over hyper-parameters, which included iteratively experimenting with various values for different hyper-parameters and also the structure of the network. The values for dropout were fixed for all dropout layers to reduce the workload of optimization, and that value was optimized over 10 equidistant steps in the range of $[0,1]$. The threshold of binary classification in the output layer of all the networks was also tuned over 10 equidistant values in the range $[0,1]$. We trained for a maximum of 150 epochs and used early stopping to get the optimal set of parameters that performed the best on the nested-validation set. We optimized for the highest micro-f1 score for all of the tuning.

\begin{table}[t]
\begin{center}
\begin{tabular}{ l | c c  }
   Model & Avg. Micro-F1 & Avg. Macro-F1\\ \hline
    CNN    & 0.493  & 0.027 \\ 
    BERT & 0.559 & 0.034 \\ 
    BERT-MeSH & 0.567  & 0.037 \\  
    \cdashline{1-3}
    LI  & 0.601 & 0.044 \\ 
    LIS & 0.599 & 0.043 \\ 
    LISA & 0.603 & 0.041 \\                                   
    LISAAS & 0.605 & 0.044 \\ 
    + Noise & \textbf{0.609}  & \textbf{0.045} \\ \hline
  \end{tabular} 
\end{center}
\caption{Average of macro-f1/micro-f1 scores for all three aspects of a clinical trial i.e. Population, Intervention/Control and Outcome. Best result for each  metric is {\bf bolded}.}
\vspace{-1em}
\label{tab:tab2}
\end{table}
\subsection{Metrics}
We evaluated our approach on 3 standard metrics (1) Precision, (2) Recall and (3) F1 score. We computed macro and micro versions of these three metrics for each of the three aspects i.e. P, I/C and O separately. Finally, we compute the average of Micro-F1 scores for all the three aspects, and this was the ultimate metric used for tuning hyper-parameters and to decide for early stopping while training.

\section{Results}
\begin{figure*}[h]
    \centering
    \includegraphics[scale=0.35]{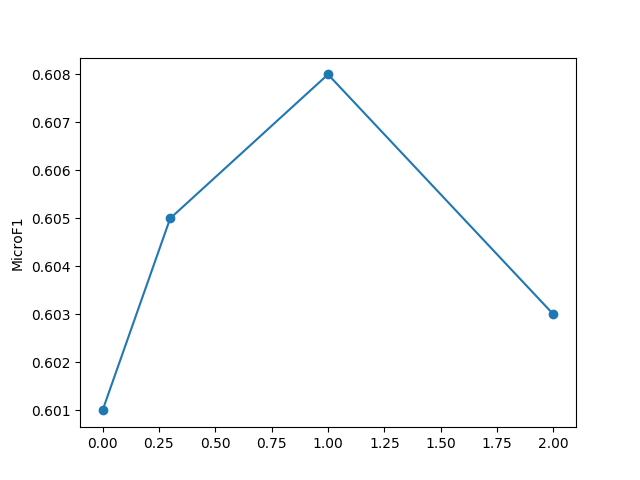}
    \includegraphics[scale=0.35]{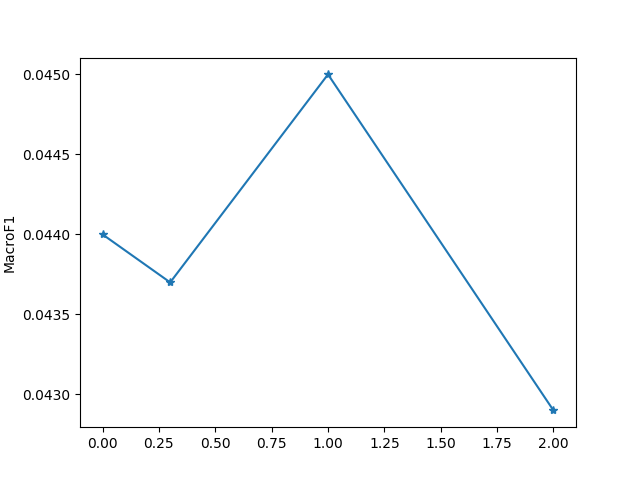}
    \caption{Performance in terms of average micro and macro-f1 scores versus the norm of the noise added to encoder representations. We scale the norm of the random noise prior to adding it. Hence, the x-axis represents the norm of the noise as a fraction of the norm of the encoder representation. }
    \label{fig:f1vsnoise}
    \vspace{-1em}
\end{figure*}

We report results for all the models using various metrics in Table \ref{tab:tab1}. CNN refers to a standard CNN-based multitask classifier that directly maps an input text into the output labels. Also, we use two separate variants of BERT, one that was trained on the BookCorpus and Wikipedia - as described in the original paper on BERT, and the other one referred to as BERT-MeSH was pre-trained on a subset of MeSH-tagged biomedical abstracts. We compare our proposed method with these competitive baselines.

 Our proposed artificial instance based approaches uniformly beat the three competitive baselines in terms of macro and micro values of precision, recall and f1 score, as can be seen in Table \ref{tab:tab1}.  This means that the model not only correctly predicts frequent labels, but also improves performance on infrequent labels.  Please note that for biomedical annotation tasks that have a large proportion of rare labels even slightly improving macro-f1 score can be very challenging, therefore, these results are encouraging.  In addition, we also compute the average of micro-f1 score and macro-f1 score for all three PICO aspects in Table \ref{tab:tab2}. Our proposed approach \textit{\textbf{LISAAS+Noise}} beats all the other baselines, especially in terms of macro metrics. Implying, the strategy of adding noise to encoder representations helps the model classify rare labels more accurately. 

Another interesting insight is that pre-training BERT on biomedical abstracts (i.e. MeSH-tagged Pubmed abstracts) performs better in comparison to simply using the original BERT pre-trained on English Wikipedia. Therefore, we can safely infer that pre-training encoders on biomedical abstracts can lead to slightly improved performance over target biomedical tasks. 

We have described the motivations for adding noise to encoder representations in the Section on Methodology. We draw the noise vector from a standard normal distribution and then scale its norm to a certain fraction of the norm of the encoder representation. In order to find out what level of noise leads to best results, we plot the average micro-f1 and macro-f1 scores for various values of this fraction. We tune this fraction over the validation set and obtain a value of 1.0, i.e. the noise has the same norm as the encoder vector representation. We were expecting the noise to be much lower for best results, but contrary to our expectations larger noise was needed by the model to handle artificial data.

\section{Conclusion}
We proposed a new method for biomedical text annotation that can work with limited training data. More specifically, our model uses pre-trained bidirectional encoders for mapping clinical texts to output concepts. Generally, such models require fine-tuning on the target task, but there is often insufficient labeled data for fine-tuning. To address this we proposed constructing extra artificial instances for fine-tuning using the labels themselves. We describe various strategies for constructing these artificial instances. Our proposed methodology leads to significantly improved results in comparison to competitive baselines. We are releasing the code\footnote{\url{https://github.com/gauravsc/pico-tagging}} used for  these experiments. 

Recently, there have been works that proposed models for generating input-text, which can be used for training text classification models. These text-generation models are trained using publically available datasets e.g. Wikipedia. Going forward, we can build upon some of these text generation methods to generate artificial abstracts for either a given biomedical concept or a group of biomedical concepts. The artificially generated abstract can then be used to train the classifier. We would investigate these research directions in the future. 

\section{Acknowledgement}
We would like to thank Cochrane for providing us with PICO annotation data.
\bibliographystyle{splncs04}
\bibliography{main}
\end{document}